\documentclass[letterpaper]{article} 
\usepackage{aaai24} 
\usepackage{times} 
\usepackage{helvet} 
\usepackage{courier} 
\usepackage[hyphens]{url} 
\usepackage{graphicx} 
\usepackage{natbib} 
\usepackage{caption} 
\usepackage{amsmath, amssymb}
\usepackage{algorithm}
\usepackage{algorithmic}
\usepackage{newfloat}
\usepackage{listings}
\DeclareCaptionStyle{ruled}{labelfont=normalfont, labelsep=colon, strut=off} 
\floatstyle{ruled}
\newfloat{listing}{tb}{lst}{}
\floatname{listing}{Listing}
\usepackage{amsfonts, array, makecell, bm, multirow, soul, cancel, xcolor, textcomp, cite, lineno}

\def \eg {\emph{e.g.}}

\title{Scale Optimization Using Evolutionary Reinforcement Learning for Object Detection on Drone Imagery}
\author{
  $^\dagger$Jialu Zhang\textsuperscript{\rm 1,3}, 
  $^\dagger$Xiaoying Yang\textsuperscript{\rm 1}, 
  Wentao He\textsuperscript{\rm 1}, 
  \thanks{Corresponding author. $^\dagger$Equal contribution.}Jianfeng Ren\textsuperscript{\rm 1,2},
  Qian Zhang\textsuperscript{\rm 1,2},
  Yitian Zhao\textsuperscript{\rm 3},
  Ruibin Bai\textsuperscript{\rm 1,2},
  Xiangjian He\textsuperscript{\rm 1,2},
  Jiang Liu\textsuperscript{\rm 3,4}
}
\affiliations{
  \textsuperscript{\rm 1}The Digital Port Technologies Lab, School of Computer Science, University of Nottingham Ningbo China\\
  \textsuperscript{\rm 2}Nottingham Ningbo China Beacons of Excellence Research and Innovation Institute, University of Nottingham Ningbo China\\
  \textsuperscript{\rm 3}Cixi Institute of Biomedical Engineering, Chinese Academy of Sciences\\
  \textsuperscript{\rm 4}Department of Computer Science and Engineering, Southern University of Science and Technology\\
  \{sgxjz1,scxxy1,scxwh1,jianfeng.ren,qian.zhang, ruibin.bai,sean.he\}@nottingham.edu.cn, yitian.zhao@nimte.ac.cn, liuj@sustech.edu.cn}

\usepackage{bibentry}

\begin{document}

\maketitle

\begin{abstract}
Object detection in aerial imagery presents a significant challenge due to large scale variations among objects. This paper proposes an evolutionary reinforcement learning agent, integrated within a coarse-to-fine object detection framework, to optimize the scale for more effective detection of objects in such images. 
Specifically, a set of patches potentially containing objects are first generated. 
A set of rewards measuring the localization accuracy, the accuracy of predicted labels, and the scale consistency among nearby patches are designed in the agent to guide the scale optimization. The proposed scale-consistency reward ensures similar scales for neighboring objects of the same category. 
Furthermore, a spatial-semantic attention mechanism is designed to exploit the spatial semantic relations between patches. 
The agent employs the proximal policy optimization strategy in conjunction with the evolutionary strategy, effectively utilizing both the current patch status and historical experience embedded in the agent.  
The proposed model is compared with state-of-the-art methods on two benchmark datasets for object detection on drone imagery. It significantly outperforms all the compared methods. 
\end{abstract}

\section{Introduction}
\label{sec: intro}

Unmanned Aerial Vehicles have been widely used in various applications, \eg, surveillance~\cite{Yun_2022_surveillance}, autonomous detection~\cite{REN_2017_UAVDetection, REN_2021_UAVDetection}, fleet navigation~\cite{Alami_2023_FleetNavigation} and agriculture~\cite{Tokekar_2016_agriculture}. 
Object detection from drone-captured images has attracted research attention recently~\cite{Xi_2021_SeanHE, Yue_2020_SeanHe, Bouguettaya_2022_review}. 
Although object detection on natural images has progressed significantly~\cite{Ge_2021_YOLOX}, detecting objects in aerial images remains challenging, mainly stemming from small scales and extreme scale variations~\cite{Deng_2021_GLSAN, Xu_2022_AdaZoom}. 

\begin{figure}[t]
	\centering
	\centerline{\includegraphics[width= 1\columnwidth]{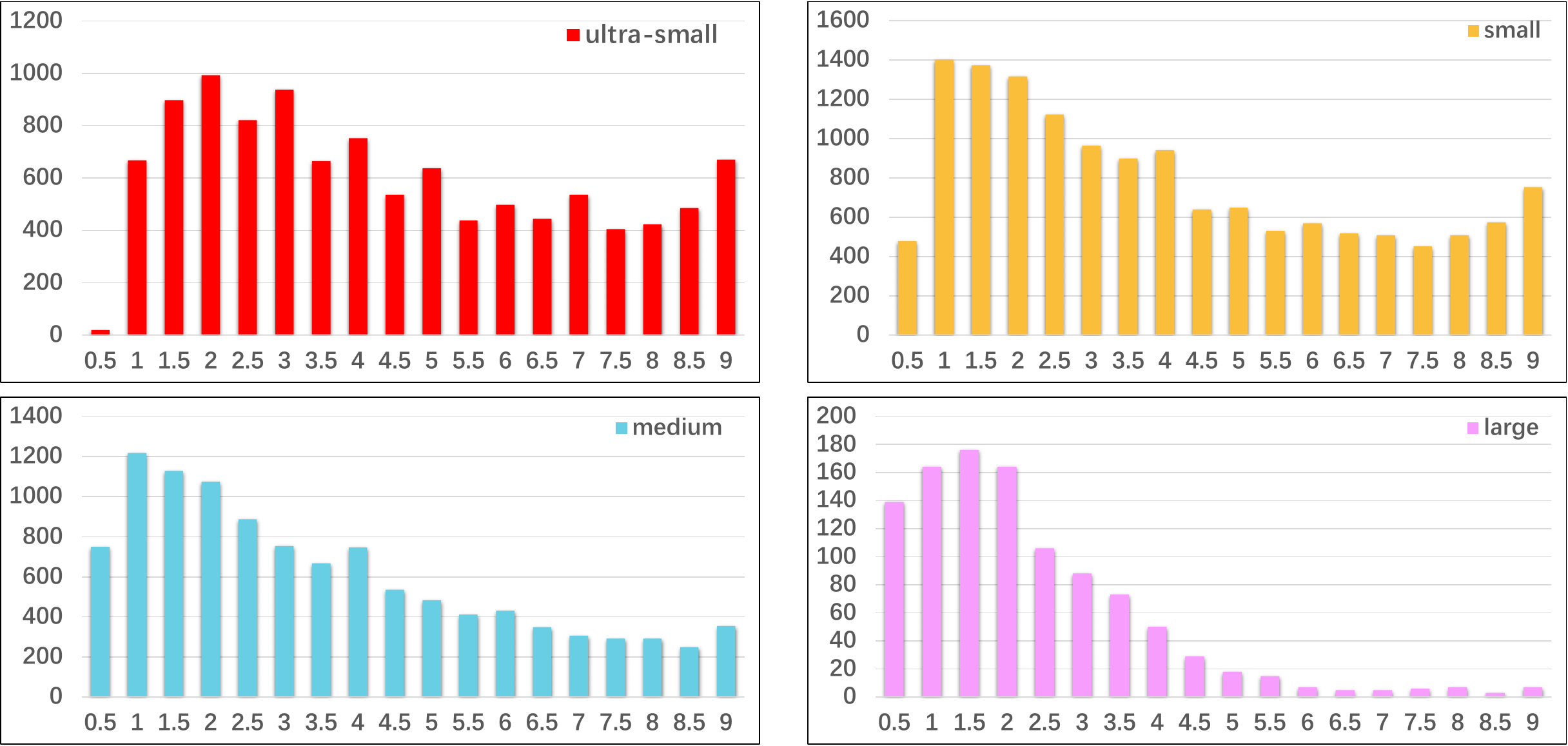}}
	\caption{
 The number of objects (\textit{y-axis}) that are optimally detected using the scaling factor (\textit{x-axis}) for ultra-small, small, medium and large objects, respectively on the VisDrone dataset. \textit{The optimal scales are significantly different for different objects.}
 }
	\label{fig: intro}
\end{figure}


Objects in aerial scenes often have large-scale variations, \eg, 
distant objects occupy few pixels while nearby objects occupy thousands. 
To tackle the challenges of detecting small objects and/or objects of different sizes, a common strategy is to divide an image into patches, scale the patches containing small objects to a fixed size~\cite{Deng_2021_GLSAN, Xu_2022_AdaZoom} or using one or more fixed scaling factors~\cite{Huang_2022_UFPMP}, and then feed them into an object detector. 
But the patch scalability is inherently limited 
due to the potential image artifacts caused by excessive scaling. 
Moreover, patches may encompass objects of different sizes. While enlarging a patch improves detecting small objects, it also enlarges large objects, potentially impeding their recognition. As shown in Fig.~\ref{fig: intro}, the optimal scales for different objects vary significantly. It is hence crucial to determine the optimal scale of each patch.

However, there lacks ground-truth annotations for the optimal scales. To tackle this problem, an EVOlutionary Reinforcement Learning (EVORL) agent is designed to determine the most suitable scale for each patch, with the guidance of a carefully designed reward function. This function assesses the image patch by considering the localization accuracy, the accuracy of predicted labels, and the scale consistency among nearby patches. The first two are directly related to the performance of object detection while the last one regularizes the optimized scales. This scale consistency stems from the inherent characteristics of drone imagery, where nearby objects of the same category tend to exhibit a similar scale. By rewarding the scale consistency, the agent is able to eliminate outliers influenced by incidental factors, thereby contributing to an improved detection performance.

Simultaneously optimizing the three rewards may result in potential conflicts, complicate the training convergence, and limit the performance. To mitigate this issue, an evolutionary strategy is integrated into the reinforcement learning framework. Specifically, the optimal scales of all patches during training are combined with sampled historical solutions to form an initial population. The proposed evolutionary algorithm refines the optimal scale determined by the current patch status by evolving the solutions using mutation and crossover, taking into account of the scale consistency among nearby patches. By incorporating both the current patch status and past experience stored in the agent's population, the proposed EVORL effectively determines the optimal scale for precise object detection.

To further boost the detection performance, a spatial-semantic attention is developed. Intuitively, spatially close objects could not only exhibit the scale consistency, but also provide the spatial-semantic attention to mutually enhance the patch features~\cite{he2023hierarchical,Zhang_2023_Spatial}. 
Specifically, the proposed method models the spatial and semantic attention by measuring the distances and the pairwise appearance correlations between adjacent objects, respectively, and aggregates these two to obtain the spatial-semantic attention. The proposed spatial-semantic attention could effectively model the spatial and semantic dependencies between objects, enhance the patch features and finally help to better detect objects at a most appropriate scale.

The proposed method follows a coarse-to-fine object detection pipeline~\cite{Bouguettaya_2022_review}. Specifically, a YOLOX~\cite{Ge_2021_YOLOX} variant is utilized to coarsely generate regions of interests. These regions are expanded to include the background context and merged to form cluster regions as in~\cite{Huang_2022_UFPMP}. A feature extractor with the proposed spatial-semantic attention is designed to visually perceive the regions. The perceived information is transmitted to the proposed EVORL agent to determine the optimal scale for each region, with the guidance of the three carefully designed rewards. Finally, the scaled regions are fed back to the detector for fine detection.

Our contributions can be summarized as follows. 1) The proposed EVORL agent is seamlessly integrated into a coarse-to-fine object detection framework, and makes use of both the current image patch and the past experience embedded in the agent to determine the optimal scale to accurately detect objects. 2) The designed reward function well addresses the challenges of lacking ground-truth labels for optimal scales, and provides supervision signals to train the agent. The proposed scale-consistency reward considers the scales of both the current object and nearby objects, to eradicate outliers and enhance the detection performance. 3) The proposed spatial-semantic attention exploits the spatial and semantic relations between nearby patches, to enhance the discriminant power of patch features. 4) The proposed method significantly outperforms state-of-the-art methods for object detection, improving the previous best average precision from 24.6\% to 28.0\% on the UAVDT dataset, and from 40.3\% to 42.2\% on the VisDrone dataset.

\section{Related Work}
\label{sec: related-work}
\subsection{Object Detection on Drone Imagery}
In aerial images, there are a large number of small objects, \eg, 26.5\% of objects in the VisDrone dataset~\cite{Zhu_2022_VisDrone} occupying fewer than $16^2$ pixels. Researchers have strove to improve small object detection on aerial imagery by adapting general object detectors on natural images. For example, \citet{Cheng_2019_Learning} designed novel objective functions for small object detection without altering existing network architectures. \citet{Li_2017_Perceptual} developed a super-resolution technique to enlarge the image for better detecting small objects. \citet{Bai_2018_SODMTGAN} utilized a generative adversarial network to obtain fine-grained features for small blurred objects. Some researchers utilized the shallow layers of deep neural networks to alleviate the problems of low resolution and detail loss caused by down-sampling operations~\cite{Bouguettaya_2022_review}, \eg, \citet{Sommer_2017_Fast} used high-resolution feature maps from earlier layers to enhance detection performance. 

Some researchers tackled the challenges of large scale variations. \citet{Wang_2019_Spatial} introduced a Receptive Field Expansion Block and a Spatial-Refinement Module to capture context information and refine solutions using multi-scale pyramid features. \citet{Zhang_2019_Scale} developed a scale-adaptive proposal network, which consists of multi-scale region proposal networks and multi-layer feature fusion to better detect objects of different scales. The feature pyramid network is often adopted to combine low-level features from shallow layers with high-level features from deep layers for multi-scale object detection~\cite{Zhou_2019_SAICFPN}.

The coarse-to-fine pipeline is often utilized for detecting objects in aerial images through extracting regions of interests using a coarse detector, scaling the image patches, and then detecting objects within them~\cite{Bouguettaya_2022_review}. \citet{Unel_2019_TilingSOD} uniformly divided the high-resolution image into patches of a fixed size, and detected objects from patches. \citet{Yang_2019_Clustered} designed a network to crop regions of dense objects and a scale estimation network to resize the crops. \citet{Xu_2022_AdaZoom} developed a self-adaptive region selection algorithm to focus on the dense regions, and leveraged super-resolution to enlarge the focused regions to a fixed size before fine-grained detection. \citet{Huang_2022_UFPMP} first equalized the scales of all generated patches, and then fed them into a unified mosaic for inference. 

Although scaling is critical to object detection, existing solutions often scale the patches to a fixed size~\cite{Xu_2022_AdaZoom} or using fixed scaling factors~\cite{Huang_2022_UFPMP}. Optimal scaling has not been fully exploited. 

\subsection{General Object Detection}
General object detectors are often adapted for drone imagery~\cite{Cai_2018_Cascade}. 
Depending on the way of feature extraction, object detectors can be broadly divided into traditional methods and deep learning methods. Traditional methods often utilize handcrafted features such as local binary patterns~\cite{Ren_2015_LBPVisualRecognition}, scale-invariant key-points~\cite{Lowe_2004_SIFT} and histograms of oriented gradients~\cite{Dalal_2005_HOG}. 
These handcrafted features are often task-specific, and ineffective in dealing with complex real-world problems~\cite{Bouguettaya_2022_review}.

\begin{figure*}[!tb]
	\centering
	\includegraphics[width= \linewidth]{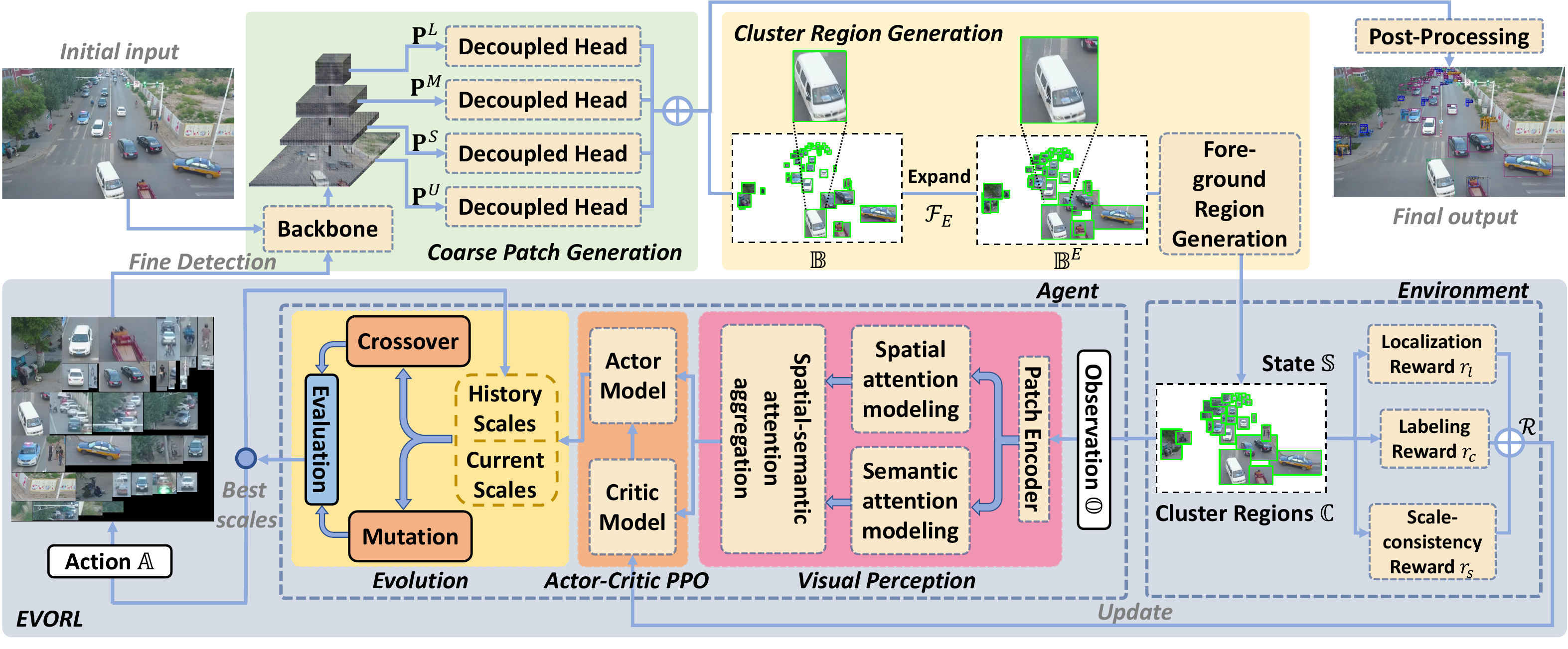}
	\caption{Overview of the proposed model.
 A YOLOX variant is first utilized to generate regions of interests. The regions are expanded to include the background context and merged to form cluster regions. An evolutionary reinforcement learning (EVORL) agent with three rewards 
 is designed to determine the optimal scale for each patch. The spatial-semantic attention is designed to boost the patch features. After determining the optimal scales through the proposed EVORL, the regions are scaled and consolidated into a mosaic image, and passed back to the detector for fine detection. 
 } 
	\label{fig: structure}
\end{figure*}

Numerous deep learning object detectors have been developed recently~\cite{Ren_2017_Faster, Ge_2021_YOLOX}, 
and they demonstrate superior performance thanks to the discriminative deep learning features. 
These models could be further categorized into two types: 1) Two-stage detectors, in which regions of interests are first extracted using a region proposal network, and objects are recognized within them. 
Representative models include 
Regions with CNN features (R-CNN)~\cite{Girshick_2014_Rich}, Faster R-CNN~\cite{Ren_2017_Faster}, and Mask R-CNN~\cite{He_2017_Mask}. 2) One-stage detectors, which integrate the proposal generation and object detection into one stage. YOLO-series (You Only Look Once)~\cite{Redmon_2016_You, Redmon_2017_YOLO9000, Ge_2021_YOLOX}, 
RetinaNet~\cite{Lin_2020_Focal} and EfficientDet~\cite{Tan_2020_EfficientDet} are the leading solutions.

General object detectors perform well on natural images, 
but not on aerial images. Aerial images often have higher image resolution but contain much more objects of various sizes, which imposes great challenges for detecting them. 

\section{Proposed Method}
\label{sec: methodology}

\subsection{Overview of Proposed Method}
\label{sec: overallFramework}



To tackle the challenges of determining the optimal scales, 
an evolutionary reinforcement learning agent is proposed. The agent is integrated into a coarse-to-fine object detection framework. The proposed framework mainly consists of three modules, as shown in Fig.~\ref{fig: structure}. 
1) \textbf{Coarse Patch Generation}. 
CSPDarkNet~\cite{Wang_2021_ScaledYOLOv4} is utilized as the backbone to generate the feature pyramid. 
In addition to the small, medium and large feature maps $\bm{\mathsf{P}}^S,\bm{\mathsf{P}}^M,\bm{\mathsf{P}}^L$ used in YOLOX~\cite{Ge_2021_YOLOX}, an ultra-small feature map $\bm{\mathsf{P}}^{U}$ is added, which contains low-level fine details for better detecting small objects. 
These features are then fed into the YOLOX decoupled heads to generate regions of interests $\mathbb{B}$. 
2) 
\textbf{Cluster Region Generation}. 
The contextual information from both the background and nearby objects has shown to be helpful in recognizing objects~\cite{
Zhang_2022_Spatial, Zhang_2023_Spatial}. The coarsely detected regions $\mathbb{B}$ are hence expanded by a factor of $\beta$ to include the background context as $\mathbb{B}^{E} = \mathcal{F}_{E}\left(\mathbb{B}; \beta \right)$,   
where $\mathcal{F}_{E}$ and $\mathbb{B}^{E}$ represent the expansion function and the expanded regions, respectively. 
The expanded regions are then clustered and merged into a cluster region set $\mathbb{C}$ using the Foreground Region Generation~\cite{Huang_2022_UFPMP}. 
3) 
\textbf{Evolutionary Reinforcement Learning}. 
A visual perception network is designed to visually perceive the regions, in which a spatial-semantic attention is designed to capture the spatial and semantic relations between nearby objects. Three rewards considering localization accuracy, label accuracy and scale consistency are designed to guide training, which well addresses the problem of lacking ground-truth annotations of optimal scales. To balance these three rewards, the hybrid algorithm combining the evolutionary strategy and Proximal Policy Optimization (PPO) strategy is designed to determine the optimal scales. 
The regions are then scaled accordingly, packed into mosaics as in~\cite{Huang_2022_UFPMP} and fed back to the detector for fine detection. Finally, post-processing techniques such as non-maximum suppression are utilized to generate the final detection results. 

\subsection{Formulation of Reinforcement Learning}
\label{sssec:problemFormulation}
The scale optimization problem is formulated as a Markov Decision Process, represented by the tuple $(\mathbb{S}, \mathbb{O}, \mathbb{A}, \mathcal{R}, p_s)$. 

\noindent\textbf{State $\mathbb{S}$} 
refers to the set of states of the environment, specifically, the determined scaling factors of all the generated cluster regions at a specific point in time. 


\noindent\textbf{Observation $\mathbb{O}$} encompasses the vital information about the objects, \eg, spatial features, semantic features, patch attributes and the attentive information from nearby objects. 



\noindent\textbf{Action $\mathbb{A} = \{a_1, \dots, a_N\}$} 
consists of a set of actions for the $N$ cluster regions, where each action $a_i$ corresponds to a specific scaling action for the cluster region $\bm{C}_i\in\mathbb{C}$. 

\noindent\textbf{State transition probability} $p_s$ is defined as $p_s(s'|s, a)= \text{Pr}\{{\mathbb{S}^{t+1}=s^\prime| \mathbb{S}^{t}=s, \mathbb{A}^{t}=a}\}$, indicating the likelihood of transitioning from the current state $s$ to a new state $s'$ under the execution of action $a$. 

\noindent\textbf{Reward $\mathcal{R}$} assesses the current state based on the object detection accuracy and the scale consistency among nearby objects. More details are provided later on.




\subsection{Visual Perception with Spatial-semantic Attention}
\label{sssec:att}

The visual perception network takes the cluster regions $\mathbb{C}$ as the input, and extracts the appearance features using a patch encoder, ResNet-18 pre-trained on ImageNet. As each region contains fewer objects than the whole image, 
the ResNet-18 can well extract the appearance features while keeping the network lightweight. 
Specifically, 
the appearance features are derived as $\bm{\mathsf{X}} = \mathcal{F}_{P}\left(\mathbb{C}; \bm{\theta}\right)$, where $\mathcal{F}_P$ represents the network, $\bm{\theta}$ represents the network parameters, and $\bm{\mathsf{X}}$ denotes all extracted features packed together. 

To capture the attentive information between nearby objects, a spatial-semantic attention is designed. Specifically, 
the spatial attention $\bm{S}$ is explicitly modeled by the reciprocal of the distance between the centers of two objects, 
where each element 
$\bm{S}_{ij} = 1/\mathcal{F}_{D}(\bm{C}_i, \bm{C}_j)$, and $\mathcal{F}_{D}$ calculates the spatial distance between $\bm{C}_i$ and $\bm{C}_j$. 
Intuitively, the smaller the spatial distance, the greater the mutual spatial attention. 




To model the semantic attention, the appearance features $\bm{\mathsf{X}}$ are firstly projected into three embedding spaces as the query matrix $\bm{Q} = \mathcal{F}_{\bm{Q}}(\bm{\mathsf{X}}, \bm{\theta_{Q}})$, key matrix $\bm{K}= \mathcal{F}_{\bm{K}}(\bm{\mathsf{X}}, \bm{\theta_{K}})$, and value matrix $\bm{V}= \mathcal{F}_{\bm{V}}(\bm{\mathsf{X}}, \bm{\theta_{V}})$, 
where $\mathcal{F}_{\bm{Q}}$, $\mathcal{F}_{\bm{K}}$ and $\mathcal{F}_{\bm{V}}$ represent the transformation networks, and $\bm{\theta_{Q}}$, $\bm{\theta_{K}}$ and $\bm{\theta_{V}}$ represent the learnable parameters of these three networks, respectively. The semantic attention is modeled as $\mathcal{F}_{S}(\bm{Q},\bm{K}) = \frac{\bm{Q} \cdot \bm{K}^\top}{\sqrt{d}}$, 
where $d$ is the feature dimension, and $\sqrt{d}$ ensures a stable gradient for the attention map. The proposed semantic attention makes use of the self-attention mechanism to exploit the attentive information between nearby objects, so that correlated objects are weighted more to boost the discriminant power of the target object.


The spatial-semantic attention $\bm{E}$ of all clustered regions is obtained through an aggregation network $\mathcal{F}_{A}(\cdot)$ 
by,
\begin{align}
	\label{eqn:FA}
  \bm{E} &= \mathcal{F}_{A}(\mathcal{F}_{S}(\bm{Q},\bm{K}) \cdot \bm{S}) \cdot \bm{V}. 
\end{align}
The proposed spatial-semantic attention well leverages on both spatial and semantic dependencies between nearby image patches, 
and hence effectively boosts the discriminative power of patch features with the support of nearby objects. 


\subsection{Reward Function}
\label{sssec:reward}

Three types of rewards are designed to provide feedback to the agent regarding the quality of a specific scaling action. 1) \textbf{Localization Reward} $r_l$, for accurately locating the objects. Specifically, 
$r_l$ calculates the average Intersection over Union (IoU) between the detected bounding boxes and the ground-truth ones, and it rewards the agent for accurately locating the objects. 
2) \textbf{Labeling Reward} $r_c$, for correctly classifying the objects. 
Specifically, $r_c$ is defined as the average classification accuracy for objects with an IoU of at least 0.5. 
3) \textbf{Scale-consistency Reward} $r_s$. 
In aerial images, 
nearby objects of the same category tend to share a similar scale. $r_s$ is designed to incentivize the scale consistency. Specifically, denote the scaling factor for $\bm{C}_i$ as $\lambda_i$. To ensure the scale consistency, for each cluster region $\bm{C}_i$, we minimize the differences between the scaling factor $\lambda_i$ and that of all its $N_i$ nearby regions of the same class, $\Delta_i = \frac{1}{N_i}\sum_{j=1}^{N_i} |\lambda_i - \lambda_j^i|$, where $\lambda_j^i$ denotes the scaling factor of the $j$-th neighboring region that has the same class label as $\bm{C}_i$.
The scale-consistency reward is defined as, 
\begin{align}\label{eq:rs}
  r_s = \frac{1}{N}\sum_{i=1}^N e^{-\Delta_i/K}, 
\end{align}
where $K$ is a normalization factor. $r_s$ is large if the neighboring cluster regions share similar scaling factors. 
Note that this reward relies on not only the optimal scaling factor of the current image patch, but also that of neighbors. Thus, the decision-making process for the optimal scaling factor of each patch becomes more complex.

The first two rewards $r_l$ and $r_c$ encourage the agent to choose a scaling factor to accurately locate and recognize the objects, and the last reward $r_s$ serves as a regularization constraint to remove the outliers in scaling factors. The reward function $\mathcal{R}$ makes use of these three rewards as, 
\begin{align}
\label{equ:rewardFunction}
\mathcal{R} = \alpha_l r_l +\alpha_c r_c + \alpha_s r_s, 
\end{align}
where $\alpha_l$, $\alpha_c$ and $\alpha_s$ are the respective weighting factors. 


\subsection{Evolutionary Reinforcement Learning Strategy}
\label{ssec:EvoRLStrategy}
The three designed rewards may conflict with each other. 
\citet{jiang2018acquisition} found that features that generated good classification scores always generated rough bounding boxes. 
Value-based Deep Q-Networks \cite{song2023siamese} or policy-based Proximal Policy Optimization (PPO) models \cite{yi2023automated} may not well address the challenges of simultaneously maximizing conflicting rewards \cite{Hui_2023_EvoRLSurvey}. Evolutionary strategies have been designed to handle conflicting rewards in multi-objective scheduling~\cite{tu2023deep,chen2022cooperative}. In this paper, an evolutionary strategy is integrated with a PPO strategy, 
where the PPO strategy effectively makes use of the appearance features 
to determine a suitable scaling action under the guidance of the three rewards, and the evolutionary strategy makes use of the past experience embedded in the agent to refine the scaling action. 

The PPO agent consists of an actor model to choose a proper action and a critic model to evaluate the action. 
Specifically, the actor model takes the spatial-semantic attended features as the input, estimates the probability distribution of feasible actions by using a squeeze-and-excitation network~\cite{hu2018squeeze}, and determines an appropriate scaling action for each cluster region. An action is sampled using the policy $\pi_\vartheta$, $a^t \sim \pi_\vartheta (a^t | s^t)$, and the advantage function is calculated to evaluate the action as $\mathcal{A}(s^t, a^t)= \mathcal{R}(s^t, a^t)+
\gamma \mathcal{V}_\varphi\left(s^{t+1}\right)-\mathcal{V}_\varphi\left(s^{t}\right)$, 
where $\gamma$ is the discount factor and 
$\mathcal{V}_\varphi(\cdot)$ is the state value 
in a specific state estimated by the critic model. 
The parameters $\vartheta$ of the actor model are updated through the gradient descent as,
\begin{align}
\label{eqn:actor_gradient}
\vartheta \leftarrow \vartheta+\eta_{\vartheta} \nabla_{\vartheta} \log \pi_\vartheta (a^t | s^t)\mathcal{A}(s^t, a^t),
\end{align}
where $\eta_{\vartheta}$ is the learning rate. 
The actor model 
performs an efficient exploration to avoid a local optimum. 
The critic model 
employs a network architecture analogous to the actor network, which takes the observations from the current state as the input and approximates the state-value function. Following the design in \cite{araslanov2019actor}, the critic loss is defined as the 
squared error loss of estimated state-value and discounted sum of rewards in the trajectory. 
The critic model is updated with a learning rate of $\eta_{\varphi}$ as,
\begin{align}
\label{eqn:critic_gradient}
\varphi \leftarrow \varphi-\eta_{\varphi}\nabla_\varphi(\mathcal{V}_\varphi\left(s^t\right)-\sum_{i=t}^T \gamma^{i-t} \mathcal{R}^i)^2. 
\end{align}

The proposed evolutionary strategy is designed to better explore and exploit the feasible action space. 
Specifically, denote $\bm{\lambda} = \{\lambda_i\}_{i=1}^{N}$ as the set of scaling factors for $N$ cluster regions. 
The scaling actions 
$\bm{\lambda}^{t}$ given by the actor model at Step $t$, along with the $W-1$ best solutions $\bm{H}^{W-1}$ from the history actions $\mathbb{H}$ form the initial population of size $W$. $\mathbb{H}$ contains effective solutions dominated by different rewards in different scenarios. By applying evolution operators such as crossover and mutation, the newly generated $W$ offspring could explore and exploit solutions in multiple scenarios, and balance the importance of different rewards. 
Specifically, the crossover of scaling factors 
combines historical solutions in different scenarios from more than one parent, and the mutation of scaling factors allows broader trials and escape from possible local optimums. 
Among $W$ parents and $W$ generated offspring, 
the new population is formed by $W$ individuals with the largest scale-consistency reward $r_s$, as defined in Eqn.~(\ref{eq:rs}). 
The evolution stops if $r_s \geq \delta$, where $\delta$ is a predefined threshold. 
The best solution after evolution is applied to scale the cluster regions, and simultaneously stored into $\mathbb{H}$. Objects are detected on the scaled regions, and the rewards are calculated to evaluate the scaling actions and update the EVORL network as in \cite{araslanov2019actor}.

The proposed evolutionary reinforcement learning for determining the optimal scales is summarized in Algo.~\ref{alg:ppo}.

\begin{algorithm}[ht]
\caption{Training procedures for the proposed EVORL}
\label{alg:ppo}
\textbf{Input}: 
The number of episodes $P$, 
the number of steps $T$, 
the number of evolution iterations $I$, 
the population size $W$

\textbf{Output}: Policy net $\pi$ 
\begin{algorithmic}[1]
\FOR{$p\gets1$ to $P$} 

  \STATE Sample a batch of $M$ images.
  
  \FOR{$t\gets 1$ to $T$}
    \STATE Derive the appearance features $\bm{\mathsf{X}}$ from images for $N$ cluster regions as $\bm{\mathsf{X}} 
   = \mathcal{F}_{P}\left(\mathbb{C}; \bm{\theta}\right)$. 
    \STATE Extract the spatial-semantic features as in Eqn.~\eqref{eqn:FA}. 
      \STATE Obtain the scaling actions $\bm{\lambda}^t$ by using the actor.
    \STATE Combine $\bm{\lambda}^t$ with $\bm{H}^{W-1}$ as the initial population. 
    \FOR{$i \gets 1$ to $I$} 
      \STATE Yield $W$ offspring by crossover and mutation.
      \STATE Evaluate each offspring and parents by Eqn.~(\ref{eq:rs}), \textbf{break} if $r_s \geq \delta$. 
      \STATE Choose best $W$ individuals as new population.
    \ENDFOR 
    \STATE Select the best $\bm{\lambda}^t$ from population and add to $\mathbb{H}$. 
    \STATE Update the state using the scaling factors $\bm{\lambda}^t$. 
    \STATE Derive the reward as $\mathcal{R}^t = \alpha_l r_l +\alpha_c r_c + \alpha_s r_s$. 
    \STATE Estimate the state-value $\mathcal{V}_\varphi\left(s^t\right)$.
    \STATE Evaluate the advantage function $\mathcal{A}(s^t, a^t)$.
    \STATE Update the actor model by using Eqn.~\eqref{eqn:actor_gradient}. 
    \STATE Update the critic model by using Eqn.~\eqref{eqn:critic_gradient}. 
  \ENDFOR
\ENDFOR
  
\end{algorithmic}
\end{algorithm}

\section{Experimental Results}
\label{sec: experiments}

\subsection{Experimental Settings}
\label{ssec: experimental settings}

\subsubsection{Datasets} 
\label{sssec:dataset}
The proposed model is compared with state-of-the-art models on two benchmark drone imagery datasets. 

\noindent \textbf{UAVDT} dataset~\cite{Du_2018_UAVDT} is a drone imagery dataset 
for object detection, single-object tracking and multi-object tracking. It contains 24,143 and 16,592 images for training and testing, respectively, with an average resolution of $1,024\times540$ pixels. 
This dataset captures images in complex scenarios and is commonly utilized for detecting vehicles like cars, trucks, and buses. 

\noindent \textbf{VisDrone} dataset~\cite{Zhu_2022_VisDrone} is a large-scale benchmark collected by drone-mounted cameras, 
encompassing 10,209 aerial images of 10 different categories, with a size of $2,000\times1,500$ pixels. The dataset is officially split into training, testing and validation sets
, with 6,471, 3,190 and 548 images, 
respectively. As ground-truth annotations of the testing set are unavailable, following 
~\cite{Liu_2021_HRDNet, Ge_2022_ZoomAndReasoning}, the validation set is used for evaluation. 

\subsubsection{Compared Methods}
The proposed method is compared against nine state-of-the-art models. The results of compared methods are taken directly from the original papers. 
Faster R-CNN~\cite{Ren_2017_Faster} serves as a baseline method. 
HRDNet adapts general object detectors on natural images for detecting small objects in aerial images~\cite{Liu_2021_HRDNet}. DMNet \cite{Li_2020_Density_Workshops} adapts the Multi-Column CNN for crowd counting to estimate an object density map and crops patches for fine detection. Other models are grouped based on the way of scaling patches in the coarse-to-fine pipeline. 

\sloppy \noindent \textbf{Resized to a fixed size:} 
SAIC-FPN utilizes super-resolution techniques to up-sample the input image and performs fine detection on cropped patches~\cite{Zhou_2019_SAICFPN}. GLSAN~\cite{Deng_2021_GLSAN} roughly detects patches first, and then resizes these patches to a fixed size by super-resolution methods. 
AdaZoom~\cite{Xu_2022_AdaZoom} leverages a reinforcement learning framework to determine the focused regions, and resizes them to a certain scale for fine detection. 

\sloppy \noindent \textbf{Resized with one or a few scaling factors:} 
ClusDet~\cite{Yang_2019_Clustered} utilizes two sub-networks, one for cropping regions of dense objects and the other for adjusting the scales of crops for fine detection. 
UFPMP-Det~\cite{Huang_2022_UFPMP} and Zoom\&Reasoning Det~\cite{Ge_2022_ZoomAndReasoning} both utilize the detector with Generalized Focal Loss \cite{Li_2020_GFL} for coarse detection. 
The former determines the patch scale by measuring the average object size inside the patch, and the latter incorporates a Foreground Zoom strategy to determine the patch scales. 


\subsubsection{Implementation Details}
\label{sssec:implementationDetails}

The stochastic gradient descent strategy is employed with a weight decay rate of 0.0005, a momentum rate of 0.9, and a dropout rate of 0.5. 
A cosine learning rate scheduler is used with an initial learning rate of 0.01. 
The same $\beta = 1.5$ is used as in~\cite{Huang_2022_UFPMP}. For the EVORL agent, the weighting factors $\alpha_l$, $\alpha_c$ and $\alpha_s$ are set to 1, the threshold $\delta = 0.5$, the size of the population $W=32$, the number of evolution iterations $I = 10$, the number of steps $T=50$ for one mini-batch, and the number of episodes $P = 1000$. 

\subsection{Comparison Results on UAVDT}
\label{sssec:uavdt}
The proposed method is compared 
to seven state-of-the-art methods on the UAVDT dataset, 
using the evaluation metrics ${AP}$, ${AP_{50}}$ and ${AP_{75}}$ as in~\cite{Huang_2022_UFPMP, Xu_2022_AdaZoom}. As shown in Table~\ref{tab:results_uavdt}, the proposed model significantly 
outperforms all previous solutions, specifically surpassing UFPMP-Det~\cite{Huang_2022_UFPMP}, the previous best performing method, by 3.4\%, 5.1\% and 3.5\% in terms of ${AP}$, ${AP_{50}}$ and ${AP_{75}}$, respectively. UFPMP-Det utilizes the average object size 
for scaling factor selection~\cite{Huang_2022_UFPMP}, which struggles with large scale variations. In contrast, the proposed EVORL makes use of both the current image patch and the past experience embedded in the agent to make 
informed decisions, adaptively determining the optimal scale for each patch. The spatial-semantic attention mechanism exploits supportive cues between objects to enhance patch features. Moreover, the Localization Reward and Labeling Reward provide supervision signals to directly maximize detection accuracy and the Scale-consistency Reward regularizes the agent to derive a more robust solution, 
leading to significant performance improvements. 

\begin{table}[!t]
 \centering
  \resizebox{\columnwidth}{!}{\begin{tabular}{lccc}
  \hline
  Method & ${AP}$ & ${AP_{50}}$ & ${AP_{75}}$ \\
  \hline
  Faster R-CNN (TPAMI, 2017) & 12.1 & 23.5 & 10.8 \\
  ClusDet (ICCV, 2019) & 13.7 & 26.5 & 12.5 \\
  DMNet (CVPR Workshop, 2020) & 14.7 & 24.6 & 16.3 \\
  GLSAN (TIP, 2021)& 17.0 & 28.1 & 18.8 \\
  AdaZoom (TMM, 2022) & 20.1 & 34.5 & 21.5 \\
  Zoom\&Reasoning Det (SPL, 2022) & 21.8 & 34.9 & 24.8 \\
  UFPMP-Det (AAAI, 2022) & 24.6 & 38.7 & 28.0 \\
  \hline
  Proposed Method & \textbf{28.0} & \textbf{43.8} & \textbf{31.5} \\  
  \hline
  \end{tabular}}%
 \caption{Comparison with the state-of-the-art methods on the UAVDT dataset. The proposed method consistently and significantly outperforms all the compared methods.} 
 \label{tab:results_uavdt}%
\end{table}%

To further analyze the performance across objects of different sizes, $AP^S$, $AP^M$ and $AP^L$, the average precision for objects with an area smaller than $32^2$ pixels, between $32^2$ and $96^2$ pixels, and larger than $96^2$ pixels, respectively on the UAVDT dataset, are summarized in Table~\ref{tab:results_sml}. 
Some methods in Table~\ref{tab:results_uavdt} did not 
report results for objects of different sizes. As shown in Table~\ref{tab:results_sml}, the proposed method consistently outperforms all the compared models across three sizes, demonstrating its capability of detecting objects of various scales. Specifically, compared 
to Zoom\&Reasoning Det~\cite{Ge_2022_ZoomAndReasoning}, 
the performance gain is 6.5\%, 7.7\% and 5.1\% for small, median and large objects, respectively. 
\begin{table}[!t]
 \centering
  \resizebox{\columnwidth}{!}{\begin{tabular}{lccc}
  \hline
  Method & ${AP^S}$ & $AP^M$ & $AP^L$ \\
  \hline
  Faster R-CNN (TPAMI, 2017) & 8.4 & 21.5 & 14.7 \\
  ClusDet (ICCV, 2019) & 9.1 & 25.1 & 31.2 \\
  DMNet (CVPR Workshop, 2020) & 9.3 & 26.2 & 35.2 \\
  AdaZoom (TMM, 2022) & 14.2 & 29.2 & 28.4 \\
  Zoom\&Reasoning Det (SPL, 2022) & 15.3 & 32.7 & 30.8 \\
  \hline
  Proposed Method & \textbf{21.8} & \textbf{40.4} & \textbf{35.9} \\
  \hline
  \end{tabular}}%
 \caption{Comparison with state-of-the-art methods on the UAVDT dataset in terms of $AP^S$, $AP^M$ and $AP^L$.}
 \label{tab:results_sml}
\end{table}

\begin{figure*}[ht]
	\centering
	\includegraphics[width=1\linewidth]{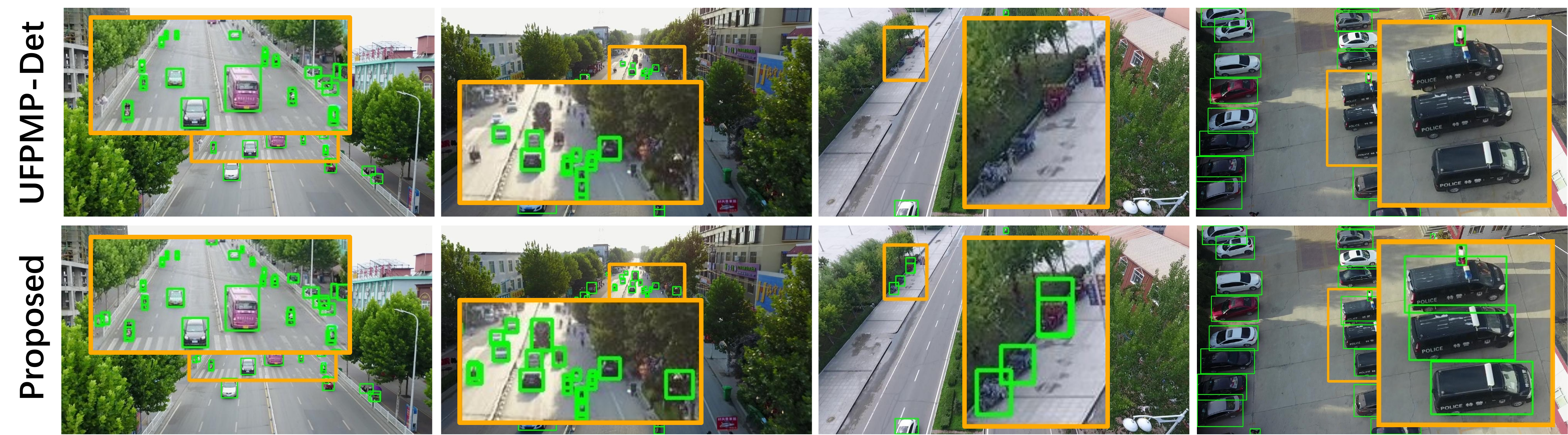}
	 \caption{Visual comparison with UFPMP-Det~\cite{Huang_2022_UFPMP} on the VisDrone dataset. 
 The proposed method correctly detects more objects than UFPMP-Det, as annotated in green. 
 }
 \label{fig: visualization}
\end{figure*}

\subsection{Comparison Results on VisDrone}
\label{sssec:visdrone}
\begin{table}[!t]
 \centering
 \resizebox{\columnwidth}{!}{
  \begin{tabular}{lccc}
  \hline
  Method & ${AP}$ & ${AP_{50}}$ & ${AP_{75}}$ \\
  \hline
  Faster R-CNN (TPAMI, 2017) & 21.8 & 41.8 & 20.1 \\
  SAIC-FPN (Neurocomputing, 2019) & 35.7 & 62.3 & 35.1 \\
  ClusDet (ICCV, 2019) & 32.4 & 56.2 & 31.6 \\
  DMNet (CVPR Workshop, 2020) & 29.4 & 49.3 & 30.6 \\
  GLSAN (TIP, 2021) & 32.5 & 55.8 & 33.0 \\
  HRDNet (ICME, 2021) & 35.5 & 62.0 & 35.1 \\
  Zoom\&Reasoning Det (SPL, 2022) & 39.0 & 66.5 & 39.7 \\
  UFPMP-Det (AAAI, 2022) & 39.2 & 65.3 & 40.2 \\
  UFPMP-Det+MS (AAAI, 2022) & 40.1 & 66.8 & 41.3 \\
  AdaZoom (TMM, 2022) & 40.3 & \textbf{66.9} & 41.8 \\
  \hline
  \multicolumn{1}{l}{Proposed Method} & \textbf{42.2} & 66.0 & \textbf{44.5} \\
  \hline
  \end{tabular}%
  }%
 \caption{Comparisons with state-of-the-art methods on the VisDrone dataset. The proposed method significantly outperforms the compared methods in terms of $AP$ and $AP_{75}$.
 }
 \label{tab:results_visdrone}%
\end{table}%

The comparison results with nine state-of-the-art methods on the VisDrone dataset~\cite{Zhu_2022_VisDrone} are summarized in Table~\ref{tab:results_visdrone}. 
Key observations are summarized as follows: 
1)~The proposed model significantly outperforms 
all compared models in terms of the key evaluation metric $AP$. Specifically, 
it achieves an $AP$ of 42.2\%, making an improvement of 1.9\% over the 
previous best model, AdaZoom~\cite{Xu_2022_AdaZoom}. AdaZoom resizes the patches to a fixed scale, while the proposed method 
utilizes the current image patch, the spatial-semantic attention, the scale consistency, and the past experience embedded in the agent to adaptively determine the most appropriate scale for each patch, 
leading to better detection performance. 2) 
The most significant performance gain is observed in $AP_{75}$, with a 2.7\% improvement over AdaZoom, thanks to the Localization Reward that enhances object localization. 
3) The proposed method yields a slightly lower $AP_{50}$ than AdaZoom, because many ultra-small objects in the VisDrone dataset only contain a few pixels, while YOLOX faces challenges in detecting these objects during coarse detection~\cite{wang2023improved}. 4) Note that 
the previous best methods on the two datasets are different. Compared to the previous best method on the VisDrone dataset, AdaZoom, the proposed method achieves significant performance gains of 7.9\%, 9.3\%, and 10.0\% in terms of $AP$, $AP_{50}$ and $AP_{75}$, respectively, on the UAVDT dataset.

\subsection{Ablation Study of Major Components}
\label{ssec: ablation-study}
The ablation results for the proposed method on the VisDrone dataset~\cite{Zhu_2022_VisDrone} are summarized in Table~\ref{tab:ablation}.
1) Compared to the YOLOX baseline, by introducing the PPO agent to determine the optimal scales based on the appearance feature extracted directly from the Patch Encoder, the $AP$, $AP_{50}$ and $AP_{75}$ are improved by 1.6\%, 2.6\% and 1.8\%, respectively. 
2) By adding the spatial-semantic attention (SSA) into the visual perception module, the $AP$, $AP_{50}$ and $AP_{75}$ are further improved by 1.6\%, 2.1\% and 1.9\%, respectively. 
3) By incorporating the evolutionary strategy into the PPO agent, the $AP$, $AP_{50}$ and $AP_{75}$ are further boosted by 1.5\%, 2.0\% and 1.5\%, respectively. 
The proposed EVORL makes good use of the past experience to refine the derived scaling factors, so that it mitigates the outlier scaling factors. These ablation results show the effectiveness of the major components in the proposed method. 

\begin{table}[!t]
 \centering
  \begin{tabular}{cccccccc}
  \hline
  YOLOX & PPO & SSA & EVO & ${AP}$ & ${AP_{50}}$ & ${AP_{75}}$ \\
  \hline
  $\surd$ & & & & 37.5 & 59.3 & 39.3 \\
  $\surd$ & $\surd$ & & & 39.1 & 61.9 & 41.1 \\
  $\surd$ & $\surd$ & $\surd$ & & 40.7 & 64.0 & 43.0 \\
  $\surd$ & $\surd$ & $\surd$ & $\surd$ & \textbf{42.2} & \textbf{66.0} & \textbf{44.5} \\
  \hline
  \end{tabular}%
 \caption{Ablation study of major components of the proposed method on the VisDrone dataset. 
 }
 \label{tab:ablation}%
\end{table}%


\subsection{Visualization of Detection Results}
\label{ssec:visualization}
The proposed method is visually compared 
to UFPMP-Det~\cite{Huang_2022_UFPMP} that yields the previous best results averaged across the two datasets. As shown in Fig.~\ref{fig: visualization}, the proposed model better recognizes small objects that are easily neglected, \eg, the `car' and `person' objects at the lower left corner of the focused regions in the first two columns, and the `tricycle' objects in the third column. The ultra-small feature map encodes more low-level but high-resolution features, partially reducing the 
loss of details during feature pooling. 
Notably, UFPMP-Det selects one of three predefined scaling factors based on the average object size in a patch, while the proposed method adaptively determines the optimal scale of each patch by utilizing both the current patch and 
the agent's past experience, and hence better detects small objects. Moreover, as seen from the last column of Fig.~\ref{fig: visualization}, UFPMP-Det wrongly classifies 
`van' as `car' whereas the proposed method can correctly classify them, thanks to the proposed scale-consistency reward and the spatial-semantic attention mechanism, which 
effectively utilizes supportive information from nearby objects to better distinguish challenging objects. 

\section{Conclusion}
\label{sec: conclusion}
To tackle the challenges of detecting small objects and handle the large scale variations in drone imagery, an evolutionary reinforcement learning framework 
has been proposed to determine the optimal scale for object detection. 
The designed agent combines the evolutionary strategy and the proximal policy optimization strategy to make good use of both the current patch status and the past experience embedded in the agent's population. The three designed rewards, considering the localization accuracy, the accuracy of predicted labels, and the scale consistency, address the issue of lacking ground-truth labels for optimal scales, and provide supervision signals for training the agent. Furthermore, a spatial-semantic attention 
has been designed to capture the mutual supportive information from nearby objects. The proposed method 
has been compared with nine state-of-the-art approaches on two benchmark datasets, UAVDT and VisDrone. It significantly outperforms the compared solutions. 

\section{Acknowledgement}
This work was supported in part by the National Natural Science Foundation of China under Grant 72071116, and in part by the Ningbo Municipal Bureau of Science and Technology under Grant 2019B10026, 2021Z089 and 2022Z173.

\bibliography{aaai24}

\end{document}